\def\year{2021}\relax
\documentclass[letterpaper]{article} 
\usepackage{aaai21}  
\usepackage{times}  
\usepackage{helvet} 
\usepackage{courier}  
\usepackage[hyphens]{url}  
\usepackage{graphicx} 
\pdfoutput=1
\usepackage{natbib} 
\usepackage{caption} 
\usepackage{xcolor}

\title{AuthNet: A Deep Learning based Authentication Mechanism using Temporal Facial Feature Movements}

\author{
    
    Mohit Raghavendra\textsuperscript{\rm 1*},
    Pravan Omprakash \textsuperscript{\rm 2*},
    Mukesh B R\textsuperscript{\rm 1},
    Sowmya Kamath\textsuperscript{\rm 1}

}
\affiliations{

    \textsuperscript{\rm 1} Department of Information Technology \\
    \textsuperscript{\rm 2} Department of Metallurgical Engineering \\
    National Institute of Technology Karnataka, India
  
    \{mohithmitra, pravanop, mukeshbr1999, \}@gmail.com

    \{sowmyakamath\}@nitk.edu.in
  
}
\begin{document}
\maketitle

\begin{abstract}
Biometric systems based on Machine learning and Deep learning are being extensively used as authentication mechanisms in resource-constrained environments like smartphones and other small computing devices. These AI-powered facial recognition mechanisms have gained enormous popularity in recent years due to their transparent,  contact-less and non-invasive nature. While they are effective to a large extent, there are ways to gain unauthorized access using photographs, masks, glasses, etc. In this paper, we propose an alternative authentication mechanism that uses both facial recognition and the unique movements of that particular face while uttering a password, that is, the temporal facial feature movements. The proposed model is not inhibited by language barriers because a user can set a password in any language. When evaluated on the standard MIRACL-VC1 dataset, the proposed model achieved an accuracy of 98.1\%, underscoring its effectiveness as an effective and robust system. The proposed method is also data efficient, since the model gave good results even when trained with only 10 positive video samples. The competence of the training of the network is also demonstrated by benchmarking the proposed system against various compounded Facial recognition and Lip reading models.
\end{abstract}

\section{Introduction}
 Biometric authentication mechanisms have been long in use in digital systems for identifying and facilitating access to restricted systems or information meant only for authorized users. Facial authentication systems have gained widespread usage in the past few years because of its contact-less non-invasive nature and ease of use in verifying a person's identity from a user's image. Early image based facial recognition systems were built on hand-engineered features such as SIFT, LBP, and Fisher vectors. Recently, highly accurate deep learning methods such as FaceNet \cite{facenet}, Baidu \cite{baidu} and DeepID models \cite{deepnet,deepId2,deepid3} have surpassed human performance. 
\begin{figure}[h!]
    \centering
    \includegraphics[width=1\columnwidth]{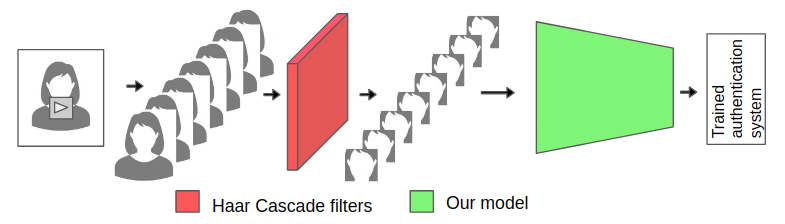}
    \caption{Data preprocessing Pipeline} 
    \label{fig:fig2}
\end{figure}
Nevertheless, facial authentication systems have often been bypassed by imposters using a photograph of the authorized user to gain access. To combat this vulnerability, models that track lip movement patterns of the user uttering a word are being explored. Deep learning models for lip reading such as LipNet \cite{DBLP:journals/corr/abs-1807-05162} have achieved a sentence-level classification accuracy of 95\%. Various other lip reading models have been built on Hidden Markov models (HMM), Long Short Term Memory (LSTM) Networks, Convolutional Neural Networks (CNN), etc. 
\begin{figure*}[ht]
    \centering
    \includegraphics[width=0.95\linewidth]{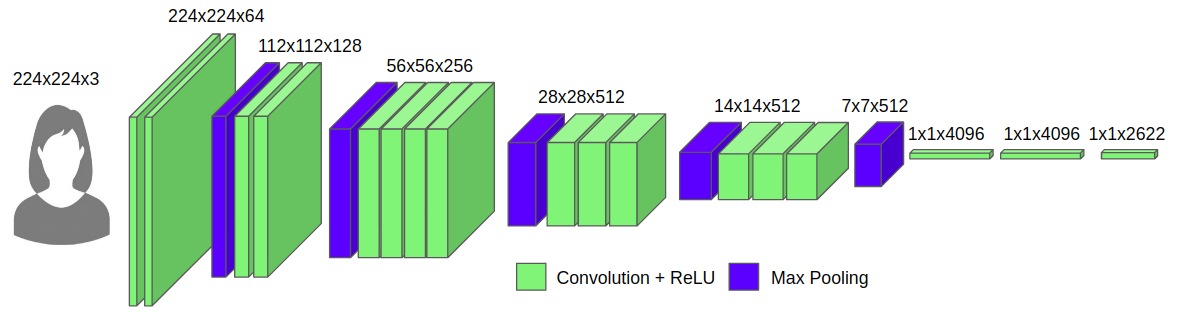}
    \caption{VGGFace Architecture (Adapted from \cite{vggface})}
    \label{fig:VGG}
\end{figure*}
However, they perform poorly in different lighting conditions and rely solely on lip movement and lip features. Deep face recognition models combined with images without the need for audio present a unique approach that are safer than basic face recognition systems and are immune to noisy environments, hence addressing the security and privacy concerns. This paper proposes an authentication model that captures the temporal facial movement patterns of a person uttering a particular word. A VGGNet model \cite{vggnet} called VGGFace \cite{vggface} pre-trained on faces was employed to capture the facial features and then the sequence of features were passed through several LSTM layers, and the model learns to predict whether a valid user is speaking their chosen password or not. The model is not provided with the actual word of the chosen password, but only the videos of its utterance. Hence, it is inherently language invariant and can be any combination of characters and numbers if spoken continuously. The proposed model was benchmarked on the publicly available standard dataset, MIRACL-VC1 \cite{dataset}, and to test the system's performance on real-world data and requirements, a separate dataset that consists of videos from smartphone cameras taken under varying conditions was compiled. The system was tested on this collated dataset as well. 

The key contribution of this paper is a strong video based facial authentication system, which uses a video of a person uttering a password, while correctly identifying the as imposter cases, which are 1) same person uttering a different word (Same Person case) or, 2) another person trying to gain access, and prevent them from gaining access ( Different person case).Another key contribution and salient point of this model is that it is language independent and domain agnostic. To our knowledge such a system has never been implemented.

The rest of this paper has been structured as follows. Section 2, discusses relevant works in the field of lip reading and authentication and the observed gaps that invite further improvements. Section 3 details the proposed approach, Section 4 describes the process by which the model was evaluated and validated, followed by conclusion and references.

\section{Related Work}

Several deep learning based face recognition models have been proposed and are popular currently \cite{facesurvey}. FaceNet \cite{facenet}, Baidu \cite{baidu} and the DeepID models \cite{deepnet,deepId2,deepid3} have shown near perfect results on benchmark datasets. These face recognition systems have been trained on millions of images and have surpassed human-level performance. However, they are still susceptible to attacks if used as authentication systems. In this section, the state-of-the-art works in this field is discussed and few observed gaps for further improvement in the performance of these systems are presented.

Mathulaprangsan \textit{et al.} \cite{lipreadsurvey} studied the various techniques used in lip reading for word classification as well as lip password authentication. Various localization, segmentation and classification approaches have been used, including conventional algorithms like FCM and LCACM that are used for lip localization in an image. They reported that the most common classifiers used for visual speech recognition at the time were, namely, Gaussian Mixture Models (GMMs) and Hidden Markov Models (HMMs). Liu \textit{et al.} \cite{lipbio} proposed a lip password verification model using Multi-boosted HMMs, focusing solely on the behavioral biometrics of lip movements. They extracted visual sequences of the mouth area region and ran them through an algorithm that segments lips in order to split the password into sub units, on which Multi-boosted HMMs were applied to formulate a decision boundary. They reported an equal error rate (EER) of 4.06\%, however, the passwords that could be used were constrained to only digits. Noda \textit{et al.} \cite{CNNlipreading} proposed a deep learning based visual speech recognition system that uses a CNN that is trained on mouth area images to predict phenomes. The outputs of the CNN were regarded as sequences and a HMM+GMM observation model was used for the word recognition task. Their work demonstrated that visual features acquired with a CNN generalized to different domains outperformed conventional approaches. 

Wand \textit{et al.} \cite{liplstm} propose a different approach for lip reading in the form of LSTM based neural networks. Sequences of mouth area images, whose features were extracted with a Histogram of Oriented Gradients (HOG) were fed into the LSTM, which was used for word classification. Their approach performed significantly better than the conventional SVM and HMM classifiers. Garg \textit{et al.} \cite{stanford} designed a new approach using a hybrid CNN and LSTM model to classify words, which was trained on the same dataset that is used in this paper, MIRACL-VC1 \cite{dataset}. They compared different approaches that included CNN+LSTM, purely CNN and SVM classifiers. They concluded that the purely CNN approach worked best for visual speech recognition attaining a 61.2\% and also noted that with more training on the CNN+LSTM model the performance might improve.

Shillingford \textit{et al.} \cite{DBLP:journals/corr/abs-1807-05162} introduced a novel approach called LipNet that performs lip reading classification on sentence level with an accuracy of 95\%. They made use of a combination of spatio-temporal convolutions and recurrent neural networks to outperform human lip readers becoming the state of the art model.

From the discussion presented on several previous works and the state-of-the-art models in the field of facial recognition based biometric systems, several vulnerabilities that need to be addressed for further improvements were observed. Facial recognition systems are not immune to attacks and vulnerabilities in the real world. Existing models for lip password verification have not achieved high accuracy, and the passwords are constrained to specific characters, for instance, only digits can be used. Moreover, most models are restricted to a certain language that they are trained on. In view of these limitations, an attempt has been made to solve these issues in this work, by designing an model that is inherently language and domain agnostic. The details of the proposed model are presented in the next section.
\\
\vspace{-1em}
\begin{figure*}[ht]
    \centering
    \includegraphics[width=0.95\linewidth]{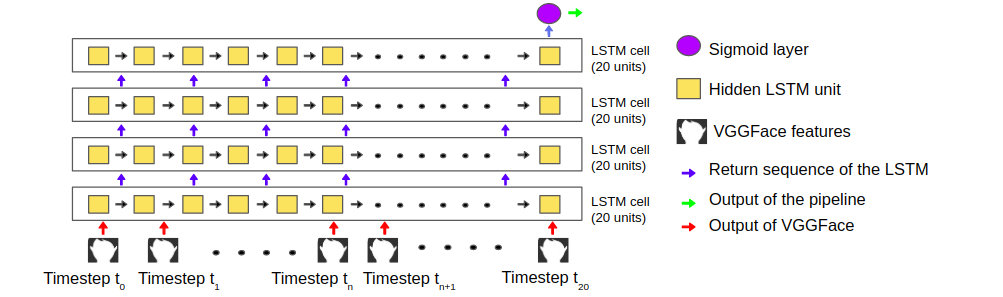}
    \caption{Proposed Architecture built on LSTM Network}
    \label{fig:LSTM}
\end{figure*}

\section{Proposed Approach}

The proposed model uses a 2 step CNN-RNN architecture depicted in Fig. \ref{fig:LSTM}. A pre-trained VGGFace model \cite{vggface} is used to extract the essential facial features at each timestep. To identify and learn the movement patterns of these features, a small network of stacked LSTM layers is employed. After the training is complete, this network can determine if a particular test case's person-word combination matches the trained person-word combination.

The MIRACL-VC1 Words dataset consists of 10 speakers speaking 10 different words, 10 times each. This has been split into different sections for different forms of testing and validation. It was randomly chosen to set aside 5 speakers i.e 5 X 10 X 10 = 500 utterances as unseen data for the model to exhibit its robustness against the Different Person Case (mentioned above). Among the remaining 5 speakers, we again set aside 3 words each, i.e 5 X 3 X 10 = 150 utterances to test the effectiveness of the model for the Same Person Case. Thus, a total of 650 examples were set aside. The remaining set has 5 X 7 X 10 = 350 utterances, in which 10 examples are positive(from the right speaker saying right word), 60 are negative samples because of the right speaker saying wrong word and 280 examples of wrong speakers. The 10 positive samples are oversampled to generate 100 new positive samples, giving a total of 350 + 100 = 450 examples. This is split into a 70 : 30 train-test split, and the the 650 examples that were set aside before are then added to this test set as negative samples,to cover all the aforementioned imposter cases.This approach is adopted to ensure that we are dealing with completely new data that model can possibly have never seen before.

In summary, for any given person-word combination, there are 10 samples of correct person correct word (oversampled to 110), 90 samples of correct person saying wrong word, 90 samples of wrong person saying correct word and 810 samples of wrong person saying wrong word, making the total of 1000 words i.e 10 speakers uttering 10 words, 10 times each.

To provide a comprehensive analysis of the performance and demonstrate the generality and robustness of the system, the model was trained in an iterative manner on all such possible combinations and tested against unseen examples. So, the training is repeated for the 5 randomly chosen speakers, each speaking 7 different words, generating 35 possible person-word combinations. 

\vspace{1em}
\noindent\textbf{Data preprocessing. }
Each training and testing sample in the MIRACL-VC1 dataset is a sequence of images of the person uttering a password, typically consisting of 5 to 15 images, as depicted in Fig. \ref{fig:fig2}. Only the color images for the words have been utilised in training. The images are first passed through a Haar cascade face detector \cite{article} to crop out the region containing faces from the images.
%(See Fig. \ref{fig:fig4}). 
The Haar cascade detector is commonly used to detect faces and face features like eyes, lips, nose, etc., and is available in the OpenCV library. It can be used to detect faces or objects in general. The algorithm works in four steps; starting with selecting Haar features, creating a set of images for easier processing, training with Adaboost and passing them through a set of cascading classifiers.
To ensure uniformity in number of timesteps, the images are padded with white images. Each image is then resized into 224 X 224 pixels to feed it into the pre-trained VGGNet model. 
For the manually compiled dataset, a similar approach was followed. This dataset represents a subset of possible real world conditions. They are first segmented into images using a constant frame rate, padded to ensure uniformity in number of timesteps and resized into 224 X 224 pixels to be fed into the VGGFace model.

% \begin{comment}
% \begin{figure}[h!]
% \centering
% \begin{tabular}{cc}
% \begin{subfigure}{0.46\linewidth}\centering\includegraphics[width=1\columnwidth]{Before.jpeg}\caption{Original Image (640x480)}\label{fig:taba}\end{subfigure}&
% \begin{subfigure}{0.46\linewidth}\centering\includegraphics[width=1\columnwidth]{After.jpeg}\caption{Resultant Image (224x224)}\label{fig:taba2}\end{subfigure}\\
% \end{tabular}
% \caption{Extracting Faces from MIRACL-VC1 dataset}
% \label{fig:fig4}
% \end{figure}
% \end{comment}

\begin{figure*}[ht]
    \centering
    \includegraphics[width=0.7\linewidth]{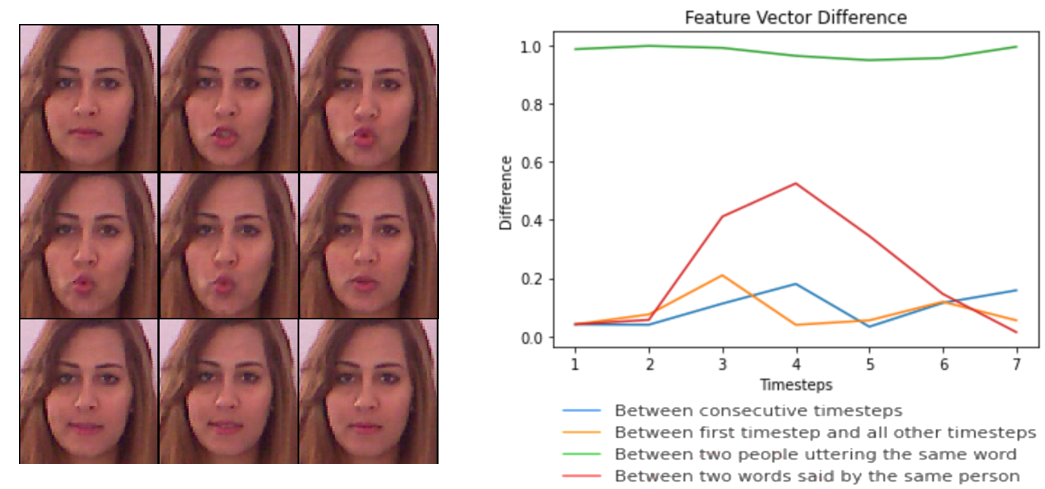}
    \caption{Variations in Features Extracted using VGGFace}
    \label{fig:extract}
\end{figure*}

\vspace{1em}
\noindent\textbf{VGGFace. }
Each image is passed through the pre-trained VGGFace model, which consists of a long sequence of convolutional layers that are trained on hundreds of thousands of images of celebrity faces. We obtain a 2,622 dimensional feature vector output for each image. This process is repeated for each word and for each person (as illustrated in Fig. \ref{fig:extract}), and reshaped to make it suitable for feeding into the LSTM layer. Thus each sample has 20 timesteps and 2622 features. The samples are labelled 1 for the correct person-word combination, and 0 for all other combinations, making this a binary classification problem. The resulting class imbalance was solved by using the technique of oversampling. 

%\vspace{-1em}
\vspace{1em}
\noindent\textbf{LSTM network. } LSTM networks have reported astounding results across a wide variety of application domains where data is sequential. So, a network of 4 LSTM layers is employed, each consisting of 20 timesteps and an output sigmoid layer for predicting probabilities. Once trained, this model can determine if the input test video contains the correct person-word combination or not, so that an authenticated user can be granted access. 

\vspace{1em}
\section{Experimental Evaluation \& Metrics}
Several experiments were performed to validate the proposed approach. The model was trained on a system with an i7-8640U CPU and 8GB of RAM. The model was bench marked on the standard MIRACL-VC1 dataset and also on a collated dataset, which consists of videos, each of 2 seconds duration. The videos were obtained from a OnePlus 6 smartphone with 1080p resolution and 16MP camera, a OnePlus 7 smartphone with 1080p resolution and 16MP camera, and an Apple iPhone XR smartphone with 720p resolution and 7MP camera. They were captured under different lighting conditions and also against varying backgrounds. The detailed statistics of each dataset used is described in Table ~\ref{datasets}.
 The proposed model took 154.2 seconds per person-word combination.
\begin{table}[h!]
\caption{Dataset statistics}
\centering
\begin{tabular}{l c c}
\hline
\textbf{Dataset}          & \textbf{MIRACL-VC1} & \textbf{Collated Dataset} \\ \hline
\# Speakers       & 10        & 3                \\ %\hline
{\# Words}          & 10        & 3                \\ %\hline
{\# Utterances}     & 10        & 6                \\ %\hline
{Image Resolution }       & $640\times480\times3$          &$350\times640\times3$ \\
{Language used}    &English &English and Hindi \\ 
\hline
\end{tabular}
\label{datasets}
\end{table}

The proposed pipeline has an binary classification task as the objective, and is designed to take a series of images of size $640 \times 480$, ordered according to timesteps, as input. The images were then passed through the pretrained VGGFace model as described in Figure \ref{fig:VGG}. The resultant feature vectors were then used for training and evaluating the LSTM network as described in Figure \ref{fig:LSTM}. The proposed model was trained using the binary crossentropy (BCE) loss function (calculated as per Eq. (\ref{BCE})) and the Adam optimizer, with an initial learning rate of 0.001 for 60 epochs and a batch size of 75 .
\begin{equation}
    \mathbf{BCE}=-{(y\log(p) + (1 - y)\log(1 - p))}
    \label{BCE}
\end{equation}

\subsection{Evaluation Metrics}
Various standard metrics were employed for assessing the performance of the proposed model, and to demonstrate its effectiveness. Sensitivity (measure as per Eq. (\ref{sensitivity})) evaluates the acceptance ratio of the system i.e., the probability that the authorised user gains rightful access. Specificity measures the rejection ratio of the system i.e., probability that an unauthorised attacker is thwarted from gaining access, given by Eq. (\ref{specificity}). Accuracy is used to measure the overall performance of the model (given by Eq. (\ref{accuracy})) and it demonstrates the efficacy of the authentication mechanism as a whole.

\begin{equation}
    % \mathbf{BCE}=-{(y\log(p) + (1 - y)\log(1 - p))}
    \mathbf{Sensitivity} = \frac{TP}{TP + FN}
    \label{sensitivity}
\end{equation}
\begin{equation}
    % \mathbf{BCE}=-{(y\log(p) + (1 - y)\log(1 - p))}
    \mathbf{Specificity} = \frac{TN}{TN + FP}
    \label{specificity}
\end{equation}
\begin{equation}
    % \mathbf{BCE}=-{(y\log(p) + (1 - y)\log(1 - p))}
    \mathbf{Accuracy} = \frac{TP + TN}{TP + TN + FP + FP}
    \label{accuracy}
\end{equation}

\vspace{0.5em}
TP and TN stand for True Positive samples, whereas FP and FN stand for False Positive and False Negative samples. Two other metrics were also used to evaluate the other aspects of the proposed model. The Equal Error Rate (EER) metric is defined as a point where the the probability of an intruder gaining access is equal to the probability that an authorised user is refused access. Hence, it is also the point where False Acceptance Ratio (FAR) is equal to the False Rejection Rate (FRR). FAR is calculated as per Eq. \ref{FAR} and FRR is calculated as per Eq. \ref{FRR}. EER is a standard metric used for comparison across biometric systems. The EER was calculated from the Receiver Operating Characteristic curve (ROC) as this is less dependent on scaling. The ROC Curve is plotted and depicted in Fig. \ref{fig:ROC}.
\begin{equation}
    % \mathbf{BCE}=-{(y\log(p) + (1 - y)\log(1 - p))}
    \mathbf{FAR} = \frac{FP}{TP + TN + FP + FN}
    \label{FAR}
\end{equation}
\begin{equation}
    % \mathbf{BCE}=-{(y\log(p) + (1 - y)\log(1 - p))}
    \mathbf{FRR} = \frac{FN}{TP + TN + FP + FN}
    \label{FRR}
\end{equation}

\vspace{0.5em}
The ROC curve is a curve of True positive rate (TPR) against False positive rate (FPR or FAR) as the threshold is varied. The ROC curve acts as a cost-benefit analysis while making decisions. It is also used to determine the threshold to ensure most true positives are gained for each false positives that the pipeline incurs.This ensures that allow as much authorised accesses are allowed while limiting the amount of unauthorised access that's allowed by the system. 
From the ROC curve intersecting with the line $ x + y = 1$ , point of equal error is found out. The Area Under the Receiver Operating Characteristic curve (AUC) is calculated  using the ROC curve. The probability that the model scores a random positive example higher than a random negative example is called AUC. AUC is a threshold-independent metric and hence measures the confidence that the model has in its decision boundary.

% \begin{equation}
%     \frac{dpH}{dt} = \frac{1}{ln(10)}(K_s(\frac{C_m}{C_s}-1)-K_a(\frac{C_m}{C_a}-1))
% \end{equation}

\subsection{Results and Discussion}
To quantify AuthNet's performance, a cross validation dataset was prepared, that accounted for the various imposter cases available for the person-word combination, namely, the  same person saying a different word, a different person saying the same word and a different person saying a different word. The different speakers chosen for testing were that of new faces completely unseen by the model during the training phase. The different words spoken by the same person were also chosen such that they are completely new and hence the capability of the model to generalize to new person-word combinations is thoroughly tested. The performance of the proposed model with other metrics methods are shown in Table~\ref{results} and the confusion matrix is shown in Fig. \ref{fig:confusion}. 

\begin{figure}[h!]
\centering
  \includegraphics[width=\columnwidth]{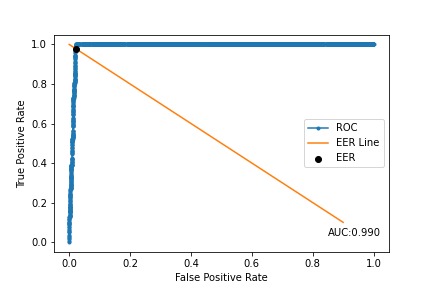}
  \caption{ROC Curve for the MIRACL-V1 dataset}
  \label{fig:ROC}
\end{figure}
\begin{figure}[h!]
  \centering
    \includegraphics[width=0.75\columnwidth]{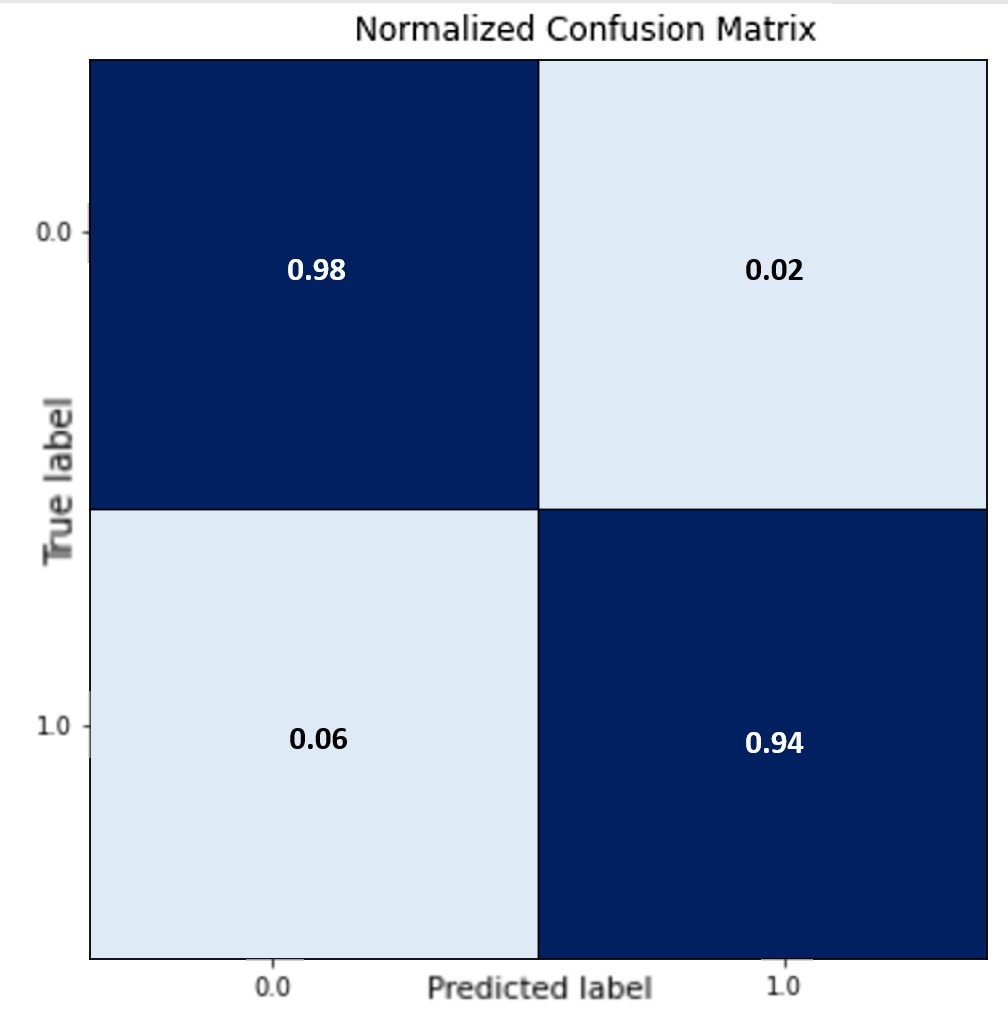}
    \caption{Confusion Matrix of proposed AuthNet model for the MIRACL-VC1 dataset}
    \label{fig:confusion}
\caption{Performance analysis of the Proposed Approach}
\label{fig:test}
\end{figure}
\begin{equation}
\cos\_diff ({\bf t},{\bf e})= 1- \frac{ \sum_{i=1}^{n}{{\bf t}_i{\bf e}_i} }{ \sqrt{\sum_{i=1}^{n}{({\bf t}_i)^2}} \sqrt{\sum_{i=1}^{n}{({\bf e}_i)^2}} }
\label{cosine}
\end{equation}
Cosine Difference, measured as per Eq. (\ref{cosine}), was used to compute the differences between images. 
The purpose of measuring cosine difference between the feature vectors across speakers and words is to show that there is a significant difference between the feature vectors obtained through the VGGFace model, among different people and even different words spoken by the same person as time progresses. This difference in feature vectors between different people and different words provides an empirical justification as to how facial features differ when people utter their passwords.Hence, when the pre-trained VGGFace model extracts a feature vector for each image, the considerable differences across images can be then be captured by the LSTM network. This is visualised in Fig. \ref{fig:extract}), This underscores the fact that the proposed model is capable of handling imposters uttering the same password, thus allowing it to be an open password system.

\begin{table}[h!]
\centering
\caption{Benchmarking the proposed AuthNet model against State-of-the-art models}
\renewcommand{\arraystretch}{1.2}
\begin{tabular}{l r}
\hline
\textbf{Models}                            & \textbf{Test Accuracy} \\ \hline

\textbf{AuthNet \textit{(Proposed)}}           & 0.980            \\ %\hline
FaceNet + Garg \textit{et al.} \\ \cite{stanford}         & 0.611            \\ %\hline
FaceNet + Chung \& Zisserman \\ \cite{Chung} & 0.652            \\ %\hline
FaceNet + Gergen \textit{et al.} \\ \cite{gergen2016dynamic}           & 0.864            \\ %\hline
FaceNet + LipNet \\ \cite{liplstm}                & 0.951           \\ \hline
\end{tabular}
\label{SOTA}
\end{table}
\vspace{-0.2em}
In Table \ref{results}, the high sensitivity value of 0.998 indicates that the system incorrectly classifies an input user-password combination 2 times out of 1000 only, to recognize the authorised user-password combination accurately. It is a good measure of AuthNet's ability to capture true positives. The high specificity value demonstrates the model's ability to correctly classify imposter attacks. The individual specificity errors for the different imposter cases are shown in Table \ref{errors}. The values show that the system denies the attacks from a different person around 98\% of the time, while it correctly identifies a wrong password being spoken 94\% of the time. This indicates the model's ability to act as an excellent defense mechanism against attacks by malicious entities, thereby ensuring a high level of privacy. From the ROC curve in Fig. \ref{fig:ROC} the point where the TPR is equal to 0.977 and FPR to be 0.023 is calculated, thus the EER value is 0.023. A low equal error rate signifies the pipeline's ability to reduce both false acceptances and false rejections to a minimum. The AUC was found to be 0.990, which signifies the pipeline's confidence in its prediction. 

On the collated dataset obtained from a capturing smartphone camera videos, we find that the performance of the model is on par with the performance on the MIRACL-VC1 dataset developed in a controlled lab setting. This signifies that the system generalises well to real world cases, even in the presence of image variance and varied lighting conditions. So, the system has an inherent capability to capture patterns unique to the user, and the results are not just a product of overfitting to the data from a simulated environment with identical background and lighting conditions.

\begin{table}[h!]
\centering
\caption{Performance of proposed AuthNet model w.r.t different datasets}
%\resizebox{9cm}{!}{
\begin{tabular}{l c c}
\hline
\textbf{Metric} &       \textbf{MIRACL-VC1} & \textbf{Collated Dataset} \\ \hline
\textbf{Sensitivity}     & 0.998             & 0.987         \\ %\hline
\textbf{Specificity}     & 0.969             & 0.976         \\ %\hline
\textbf{Accuracy}        & 0.980             & 0.981         \\ %\hline
\textbf{AUC Score}       & 0.990             & 0.989         \\ %\hline
\textbf{Equal Error Rate}& 0.023             & 0.037         \\ \hline
\end{tabular}
%}
\label{results}
\end{table}

To demonstrate the effectiveness of the training, AuthNet model is benchmarked against a two-level system -  several state-of-the-art models in face recognition combined with lip reading. Since FaceNet \cite{facenet} is the current best-performing face recognition model reporting an accuracy of 99.76\%, it was used for comparison. Garg \textit{et al. }\cite{stanford} proposed a lip reading model that achieved an accuracy of 61.2\%. The models of Chung \& Zisserman \cite{Chung} and Gergen \textit{et al.} \cite{gergen2016dynamic} were the previous state-of-the-art models in lip reading on word level with their reported accuracy values of 65.4\% and 86.6\% respectively. Lipnet \cite{DBLP:journals/corr/abs-1807-05162}  has been included to benchmark the performance of the proposed AuthNet model. The scores and comparisons are reported in Table \ref{SOTA}, which shows that the performance of the  trained model matches the combined performance of state-of-the-art models while overcoming errors or imposter attacks in these models.

\begin{table}[h!]
\centering
\caption{Specificity for different types of errors (MIRACL-VC1 dataset)}
%\resizebox{8cm}{!}{
\setlength{\tabcolsep}{4pt}
\renewcommand{\arraystretch}{1.2}
\begin{tabular}{p{4.69cm} c c}
\hline
\textbf{Imposter types/Error cases} &  \textbf{Samples} & \textbf{Specificity} \\ \hline
Same Person-Different Words       & 210   & 0.9420               \\ %\hline
Different Person-Same Words        & 150   & 0.9893            \\ %\hline
Different Person-Different Words  & 100   & 0.9730         \\ \hline
\end{tabular}
%}
\label{errors}
\end{table}

\section{Conclusion and Future Work}
In this paper, AuthNet, a temporal facial movement based authentication mechanism using 2-level trained CNN-RNN deep neural models was presented. When evaluated on the standard MIRACL-VC1 dataset, AuthNet achieved an accuracy of 98.1\%, underscoring its effectiveness. The results obtained from cross-validation on a collated dataset proved its inherent language and domain agnostic nature. It also demonstrated the real-world application of such a system in mobile phones and other resource constrained smart devices, involving varying backgrounds, different lighting contents, varying video resolutions, etc. 

AuthNet can function as an open password system, as the model effectively classified the same password being spoken by a different person. Hence, disclosing the password does not cause any security concerns. It also has no language barrier, since the model is not dependent on knowledge of the language in which the password/phrase is being spoken, but only the temporal facial movements of the speaker extracted from the videos of its utterance. The proposed method is also data efficient, since the model gave good results even when trained with only 10 positive video samples. As part of future work, the proposed model could be optimised in its time complexity, as it takes approximately 2 seconds for testing per sample currently. Since fast processing is a critical requirement for deployment as an authentication system for smartphones and computers, optimising the testing time will be a primary concern to be addressed, which will be explored.

\bibliography{article.bib, baidu.bib, chung.bib, CNNlipreading.bib, dataset.bib, DBLPjournalscorrabs-1807-05162.bib, deepId2.bib, deepid3.bib, deepnet.bib,facenet.bib, facesurvey.bib, gergen2016dynamic.bib, lipbio.bib, liplstm.bib, lipreadsurvey.bib, stanford.bib,vggface.bib, vggnet.bib}

\begin{thebibliography}{18}
\providecommand{\natexlab}[1]{#1}
\providecommand{\url}[1]{\texttt{#1}}
\providecommand{\urlprefix}{URL }
\expandafter\ifx\csname urlstyle\endcsname\relax
  \providecommand{\doi}[1]{doi:\discretionary{}{}{}#1}\else
  \providecommand{\doi}{doi:\discretionary{}{}{}\begingroup
  \urlstyle{rm}\Url}\fi

\bibitem[{amit, jnoyola, and sameepb(2016)}]{stanford}
amit, A.~G.; jnoyola, J.~N.; and sameepb, S.~B. 2016.
\newblock Lip reading using CNN and LSTM.

\bibitem[{Balaban(2015)}]{facesurvey}
Balaban, S. 2015.
\newblock Deep learning and face recognition: the state of the art.
\newblock 94570B.
\newblock \doi{10.1117/12.2181526}.

\bibitem[{Chung et~al.(2017)Chung, Senior, Vinyals, and Zisserman}]{Chung}
Chung, J.~S.; Senior, A.; Vinyals, O.; and Zisserman, A. 2017.
\newblock Lip Reading Sentences in the Wild.
\newblock 3444--3453.
\newblock \doi{10.1109/CVPR.2017.367}.

\bibitem[{Gergen et~al.(2016)Gergen, Zeiler, Hussen~Abdelaziz, Nickel, and
  Kolossa}]{gergen2016dynamic}
Gergen, S.; Zeiler, S.; Hussen~Abdelaziz, A.; Nickel, R.; and Kolossa, D. 2016.
\newblock Dynamic Stream Weighting for Turbo-Decoding-Based Audiovisual ASR.
\newblock 2135--2139.
\newblock \doi{10.21437/Interspeech.2016-166}.

\bibitem[{Liu et~al.(2015)Liu, Deng, Bai, and Huang}]{baidu}
Liu, J.; Deng, Y.; Bai, T.; and Huang, C. 2015.
\newblock Targeting Ultimate Accuracy: Face Recognition via Deep Embedding .

\bibitem[{Liu and Deng(2015)}]{vggnet}
Liu, S.; and Deng, W. 2015.
\newblock Very deep convolutional neural network based image classification
  using small training sample size.
\newblock 730--734.
\newblock \doi{10.1109/ACPR.2015.7486599}.

\bibitem[{Liu and Cheung(2014)}]{lipbio}
Liu, X.; and Cheung, Y.-m. 2014.
\newblock Learning Multi-Boosted HMMs for Lip-Password Based Speaker
  Verification.
\newblock \emph{Information Forensics and Security, IEEE Transactions on} 9:
  233--246.
\newblock \doi{10.1109/TIFS.2013.2293025}.

\bibitem[{Mathulaprangsan et~al.(2015)Mathulaprangsan, Wang, Frisky, Tai, and
  Wang}]{lipreadsurvey}
Mathulaprangsan, S.; Wang, C.-Y.; Frisky, A.; Tai, T.-C.; and Wang, J.-C. 2015.
\newblock A survey of visual lip reading and lip-password verification.
\newblock 22--25.
\newblock \doi{10.1109/ICOT.2015.7498485}.

\bibitem[{Ouyang et~al.(2014)Ouyang, Luo, Zeng, Qiu, Tian, Li, Yang, Wang,
  Xiong, Qian, Zhu, Wang, Loy, Wang, and Tang}]{deepnet}
Ouyang, W.; Luo, P.; Zeng, X.; Qiu, S.; Tian, Y.; Li, H.; Yang, S.; Wang, Z.;
  Xiong, Y.; Qian, C.; Zhu, Z.; Wang, R.; Loy, C.~C.; Wang, X.; and Tang, X.
  2014.
\newblock DeepID-Net: Deformable Deep Convolutional Neural Networks for Object
  Detection \doi{10.1109/CVPR.2015.7298854}.

\bibitem[{Parkhi, Vedaldi, and Zisserman(2015)}]{vggface}
Parkhi, O.; Vedaldi, A.; and Zisserman, A. 2015.
\newblock Deep Face Recognition.
\newblock volume~1, 41.1--41.12.
\newblock \doi{10.5244/C.29.41}.

\bibitem[{Rekik, Ben-Hamadou, and Mahdi(2014)}]{dataset}
Rekik, A.; Ben-Hamadou, A.; and Mahdi, W. 2014.
\newblock A New Visual Speech Recognition Approach for RGB-D Cameras.
\newblock ISBN 978-3-319-11754-6.
\newblock \doi{10.1007/978-3-319-11755-3_3}.

\bibitem[{Schroff, Kalenichenko, and Philbin(2015)}]{facenet}
Schroff, F.; Kalenichenko, D.; and Philbin, J. 2015.
\newblock FaceNet: A unified embedding for face recognition and clustering.
\newblock 815--823.
\newblock \doi{10.1109/CVPR.2015.7298682}.

\bibitem[{Shillingford et~al.(2018)Shillingford, Assael, Hoffman, Paine,
  Hughes, Prabhu, Liao, Sak, Rao, Bennett, Mulville, Coppin, Laurie, Senior,
  and Freitas}]{DBLP:journals/corr/abs-1807-05162}
Shillingford, B.; Assael, Y.; Hoffman, M.; Paine, T.; Hughes, C.; Prabhu, U.;
  Liao, H.; Sak, H.; Rao, K.; Bennett, L.; Mulville, M.; Coppin, B.; Laurie,
  B.; Senior, A.; and Freitas, N. 2018.
\newblock Large-Scale Visual Speech Recognition.

\bibitem[{Sun et~al.(2015)Sun, Liang, Wang, and Tang}]{deepid3}
Sun, Y.; Liang, D.; Wang, X.; and Tang, X. 2015.
\newblock DeepID3: Face Recognition with Very Deep Neural Networks .

\bibitem[{Sun, Wang, and Tang(2014)}]{deepId2}
Sun, Y.; Wang, X.; and Tang, X. 2014.
\newblock Deep Learning Face Representation by Joint
  Identification-Verification.
\newblock \emph{Proc. NIPS} 27.

\bibitem[{Viola and Jones(2004)}]{article}
Viola, P.; and Jones, M. 2004.
\newblock Robust Real-Time Face Detection.
\newblock \emph{International Journal of Computer Vision} 57: 137--154.
\newblock \doi{10.1023/B:VISI.0000013087.49260.fb}.

\bibitem[{Wand, Koutnik, and Schmidhuber(2016)}]{liplstm}
Wand, M.; Koutnik, J.; and Schmidhuber, J. 2016.
\newblock Lipreading with long short-term memory.
\newblock 6115--6119.
\newblock \doi{10.1109/ICASSP.2016.7472852}.

\bibitem[{Özcan and Basturk(2019)}]{CNNlipreading}
Özcan, T.; and Basturk, A. 2019.
\newblock Lip Reading Using Convolutional Neural Networks with and without
  Pre-Trained Models.
\newblock \emph{Balkan Journal of Electrical and Computer Engineering}
  195--201.
\newblock \doi{10.17694/bajece.479891}.

\end{thebibliography}
\end{document}